\documentclass[a4paper]{svproc}
\usepackage{pdflscape}

\usepackage{algorithmic}
\usepackage{graphicx}
\usepackage{rotating}
\usepackage{comment}
\usepackage{amsmath}
\usepackage{subfig}
\usepackage[utf8]{inputenc}
\usepackage{amssymb}
\usepackage[ruled,linesnumbered]{algorithm2e}
\usepackage{amsfonts,amssymb,amsmath,amsgen,amsopn,amsbsy,theorem,graphicx,epsfig}
\usepackage{amscd,bezier,latexsym,mathrsfs,enumerate}
\usepackage[utf8]{inputenc}
\usepackage[english]{babel}

\usepackage{tabularx, booktabs}
\usepackage{cite}
\usepackage[table]{xcolor}

\usepackage{flushend}
\usepackage{hyperref}

\hypersetup{
    colorlinks=true,
    linkcolor=magenta,      % internal links
    filecolor=magenta,      % links to local files
    urlcolor=magenta        % external links
}

\begin{document}
\mainmatter              % start of a contribution
\title{C$^2$DA: Contrastive and Context-aware Domain Adaptive Semantic Segmentation}
\titlerunning{Domain Adaptive Segmentation}  % abbreviated title (for running head)
\author{Md. Al-Masrur Khan\inst{1} \and Zheng Chen\inst{1}
\and Lantao Liu \inst{1} }
\authorrunning{Khan et al.} % abbreviated author list (for running head)
\tocauthor{Md. Al-Masrur Khan, Zheng Chen and Lantao Liu}
\institute{Luddy School of Informatics, Computing,
and Engineering, \\Indiana University, Bloomington, IN 47408, USA.\\
\email\{khanmdal, zc11, lantao\}@iu.edu\\ %WWW home page:
%\texttt{https://vail.sice.indiana.edu/}
}

\maketitle              % typeset the title of the contribution
\vspace{-15pt}
\begin{abstract}
Unsupervised domain adaptive semantic segmentation (UDA-SS) aims to train a model on the source domain data (e.g., synthetic) and adapt the model to predict target domain data (e.g. real-world) without accessing target annotation data. Most existing UDA-SS methods only focus on inter-domain knowledge to mitigate the data-shift problem. However, learning the inherent structure of the images and exploring the intrinsic pixel distribution of both domains are ignored; which prevents the UDA-SS methods from producing satisfactory performance like the supervised learning. Moreover, incorporating contextual knowledge is also often overlooked. Considering these issues, in this work, we propose a UDA-SS framework that learns both intra-domain and context-aware knowledge. To learn the intra-domain knowledge, we incorporate contrastive loss in both domains, which pulls pixels of similar classes together and pushes the rest away, facilitating intra-image-pixel-wise correlations. To learn context-aware knowledge, we modify the mixing technique by leveraging contextual dependency among the classes to learn context-aware knowledge. Moreover, we adapt the Mask Image Modeling (MIM) technique to properly use context clues for robust visual recognition, using limited information about the masked images. Comprehensive experiments validate that our proposed method improves the state-of-the-art UDA-SS methods by a margin of 0.51$\%$ mIoU and 0.54$\%$ mIoU in the adaptation of GTA-V$\rightarrow$Cityscapes and Synthia$\rightarrow$Cityscapes, respectively. We open-source our C$^2$DA code. Code link: \href{https://github.com/Masrur02/C-Squared-DA}{github.com/Masrur02/C-Squared-DA}
\end{abstract}
\vspace{-30pt}
\section{INTRODUCTION}
% Semantic segmentation is a pivotal area of deep learning, where the classification is performed at a pixel level. Semantic segmentation plays an indispensable role across several fields such as robotic perception \cite{richter2016playing}, biomedical classification \cite{ronneberger2015u}, and numerous industrial applications \cite{rubio2019multi}. Currently, the mainstream semantic segmentation methods \cite{xie2021segformer}, \cite{zhang2021k}, \cite{strudel2021segmenter} are based on deep neural networks which are learned in a supervised manner. A big limitation of supervised learning is the need for labeled data. Literature reveals that labeling for a single image to use on a semantic segmentation task may take more than 1.5 hours \cite{chen2023ida}. Though there are a few numbers of datasets including Cityscapes \cite{cordts2016cityscapes}, ACDC \cite{sakaridis2021acdc}, Synthia \cite{ros2016synthia}, KITTI \cite{geiger2013vision} and so on, 
Deep segmentation models trained by existing datasets \cite{cordts2016cityscapes, sakaridis2021acdc, ros2016synthia} are not always sufficient to segment accurately in novel and difficult environments. Most importantly when it comes to the case of different domains, for example, simulation-real, day-night, summer-fall, and so on, it is difficult to boost the model to be generalized in unseen data of other domains because there is a \textit{non-trivial} domain shift between the trained data and the data to be predicted. Unsupervised Domain Adaptation (UDA) is an effective framework for solving the problem of limited annotated data and the problem of domain shift in semantic segmentation. In UDA, a model is trained to transfer the knowledge in the annotated data (source domain) to non-annotated data (target domain). The source and target domain pair can be simulation-real, day-night, summer-fall, etc. Current UDA methods have shown mesmerizing performance; however, they still have not reached the level of accuracy compared to that of supervised learning. Over the years, researchers have resorted to various ideas including adversarial learning \cite{gong2021dlow}, %\cite{hoffman2018cycada}, 
self-training \cite{hoyer2022daformer}, %\cite{hoyer2022hrda}, 
consistency regularization \cite{sajjadi2016regularization}, %\cite{sohn2020fixmatch} 
to minimize the discrepancy between the data distribution of the source domain and the target domain. However, still, a couple of issues are mostly overlooked in UDA frameworks. Existing works usually focus on either only intra-domain class-wise knowledge \cite{chen2023pipa} or only context-aware knowledge~\cite{hoyer2023mic}. 
\vspace{-15pt}
\begin{figure}[h]
{
    \centering
    \includegraphics[width=1\linewidth]{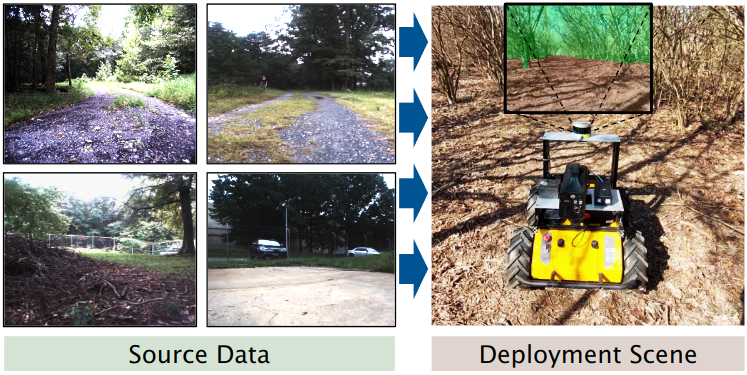} \vspace{-10pt}
    \caption{\small Robots are usually \textit{deployed} to an environment that has a non-trivial \textbf{domain shift} from the \textit{source} data, where the human's label/knowledge is provided. Our work is to support this kind of transfer learning tasks. 
    \vspace{-5pt} 
    \label{title} 
}
}
\end{figure}
\vspace{-15pt}

In this work, 
% On one hand, while focusing on inter-domain knowledge, adapting intra-domain (class-wise) knowledge is often ignored; which degrades the performance of UDA \cite{chen2023pipa}. On the other hand, considering context-aware knowledge is also overlooked during adaptation from the source domain data to target domain data. Considering context-aware knowledge can reduce the confusion between classes with similar visual appearances \cite{hoyer2023mic} as well as ensure realistic spatial distribution among the classes during cross-domain mixing \cite{zhou2022context}. 
% Considering the aforementioned observation, 
we propose a unified UDA framework that tightly couples the intra-domain knowledge and the context-aware knowledge. To learn the intra-domain knowledge, we explore pixel-to-pixel relationships to understand the inherent structures of intra-domain images. This approach ensures intra-class compactness as well as inter-class separability. By mapping the pixels into an embedding space, discriminative feature learning can be obtained. This is achieved by pulling together pixels belonging to the same class and pushing apart pixels from different classes, thereby promoting both intra-class compactness and inter-class separability. To adapt the contextual knowledge, our strategy is twofold. First, we adopt a mixing technique informed by the prior knowledge that the source and target domains usually share associative semantic contexts (e.g., a rider is always positioned above a bicycle or motorcycle). Therefore, we modify the ClassMix technique \cite{olsson2021classmix} by leveraging accumulated spatial distributions and providing semantically aware information, yielding important clues of context dependency. Second, we incorporate the Mask Image Modeling (MIM) technique to help the model understand the target domain's context relations during UDA. We mask out random patches from the target domain data and train the model to infer the segmentation map of the entire image along with the masked-out parts. In this way, the model is compelled to utilize the contexts to identify the masked-out areas. We mask out different patches throughout the training, ensuring robust learning of the context clues. Moreover, in this work, we do not confine the proposed framework to only adapting in urban scenes rather we perform the domain adaptation in forest scenes as well. 

Finally, we deploy our UDA framework in a robotic vehicle to navigate it in a forest environment. The necessity of applying domain adaptation in robotics is strong because robots can be deployed in a wide range of environments. Fig. \ref{title} shows an example of the difference between the training source data and target deployment data. The available training data only contains images collected from a summer forest, while the deployment task desires the robot to operate in a fall forest that exhibits a drastically changed visual appearance, thus leading to a non-trivial domain shift.
% In summary, our contributions to this work are as follows:
% \begin{itemize}
% \item We propose a unified UDA framework that considers intra-domain and context-aware knowledge during UDA.
% \item  We extend the application of UDA for solving the domain adaptation in a forest environment
% \item We deploy our UDA framework in a robotic vehicle to navigate it in a forest environment.
% \end{itemize}
Our comprehensive evaluations across well-known datasets reveal that our UDA framework outperforms the current state-of-the-art (SOTA) models under the same settings.
\vspace{-10pt}
\section{RELATED WORKS}
\vspace{-5pt}
\textbf{Unsupervised Domain Adaptation:}
In UDA a deep learning model is trained on annotated source data to predict on label-scarce target domain data.  Over the years, UDA has been employed in almost all branches of computer vision including classification \cite{ganin2016domain}, segmentation \cite{hoyer2022daformer}, object detection \cite{chen2021scale} due to its ability to solve domain gaps. Mainstream UDA methods are mainly grouped into two categories: adversarial training \cite{hoffman2018cycada} and self-training \cite{hoyer2022hrda}. Adversarial training focuses on learning domain-invariant knowledge through the alignment of features from the source domain with those of the target domain. It mitigates the domain gaps by using entropy minimization \cite{long2016unsupervised}, correlation alignment \cite{sun2016deep}, etc. Adversarial learning focuses on aligning features between two domains on a global scale rather than aligning features at the class level and leads to a negative transfer problem during semantic segmentation tasks. As a result, the training process becomes unstable and yields suboptimal performance. On the other hand, self-training adapts a student-teacher \cite{hoyer2022daformer} framework to tackle the data shift problem which is typical in UDA. In this strategy, pseudo-labels are generated by a teacher model trained on the source domain data. Later, the student model is trained on the target images by leveraging those pseudo-labels as ground truth data. %Recently, self-training methods have been widely adopted in UDA tasks for their groundbreaking adaptation using pseudo labels. 
However, due to significant differences in data distributions between the two domains, pseudo-labels inherently possess noise. To create reliable pseudo-labels, strategies utilized often include adopting pseudo labels with high confidence \cite{zou2018unsupervised}, learning from the future \cite{du2022learning}, employing uncertainty-aware pseudo labeling \cite{zheng2021rectifying}, and so on. Other applied strategies to increase the robustness of UDA methods are utilizing consistency regularization \cite{sohn2020fixmatch}, domain-mixup \cite{zhou2022context}, multi-resolution inputs \cite{hoyer2022hrda}, etc.

% However, most of the UDA frameworks ignore intra-domain knowledge and context-aware knowledge for mitigating the domain shift between the domains.   \textcolor {red} {I am afraid that, this is not our novel contribution. Several works used contrastive learning and context-aware learning in UDA. We might need to produce some statements that will clearly state our contribution and the difference from previous works. }

\textbf{Contrastive Learning:}
Contrastive learning is one of the notable methods for learning discriminative feature representations. It learns the inherent structure of the images by contrasting positive data pairs against the negative data pairs. For recognizing positive and negative pairs, computing Euclidean distance  \cite{chopra2005learning} and cosine similarity  \cite{zhao2021contrastive} between features are the widely adopted methods. Along with the classification, contrastive learning has also been utilized in semantic segmentation tasks \cite{hu2021region}. As semantic segmentation is a pixel-level classification task, pixels from the same classes are considered as positive pairs and the rest are considered as negative pairs. So, contrastive learning in semantic segmentation tries to pull the pixels of the same classes together while pushing away the rest of the pixels. In addition, to pixel-wise contrastive learning, recent studies have also explored other forms of contrastive learning for segmentation tasks, such as prototype-wise  \cite{hu2021region} and distribution-wise \cite{li2021semantic} approaches. Contrastive learning can be performed in both supervised \cite {khosla2020supervised} and self-supervised \cite{he2020momentum} manners. Recently UDA methods are also exploring contrastive loss for its ability to learn the feature representations \cite{vayyat2022cluda}. 
% \textcolor {red} {Other works are also using contrastive learning. What is the difference between our setup? I need help. Do we need a literature review on context-aware learning also? }

\textbf{Masked-based Learning:}
Masked language modeling has reshaped the field of natural language processing (NLP) \cite{brown2020language} by masking the tokens of input sequences. Recently, masking-based learning has demonstrated itself in the name of Masked Image Modeling (MIM) as a competitive challenger for self-supervised learning in computer vision \cite{bao2021beit, dosovitskiy2020image}. In MIM, the models are expected to reconstruct various features including VAE features \cite{li2022mc}, HOG features \cite{wei2022masked}, or color information \cite{he2022masked} of the masked image patches. For masking the patches researchers have explored different techniques, for example, random-patch masking \cite{wei2022masked}, block-wise masking \cite{bao2021beit}, attention-guided masking \cite{kakogeorgiou2022hide}, etc. Starting from the context encoder approach \cite{pathak2016context}, MIM has also been explored on the modern vision transformers \cite{dosovitskiy2020image, bao2021beit}. Most of the MIM works are based on transformer-decoder and reconstruction techniques. Relevant to these works, a recent study SimMiM \cite{xie2022simmim} utilized a linear head-based approach, and \cite{xuestare} performed MIM without reconstruction however. Recently, MIM has also been explored for UDA tasks in \cite{hoyer2023mic} for its capability to learn contextual relations.

%In addition to the transformer decoder-based approach, recent studies have %also explored other techniques such as the linear head-based approach %\cite{xie2022simmim}, or 
\vspace{-10pt}
\section{Methodology}
\vspace{-5pt}
To explain the overall method of our work, we first provide preliminary knowledge of the UDA methods in Sect. \ref{sec:preliminary}. Then we provide the general structure of our proposed model in Sect. \ref{sec:framework}. In Sect. \ref{sec:masking}, we discuss the technique of generating masked images and learning from context clues. In Sect. \ref{sec:contrastive}, we discuss the technique for obtaining intra-domain knowledge. Finally, in Sect. \ref{sec:context} we describe the context-aware mixing scheme for reducing the domain shift.
\vspace{-12pt}
\subsection{Preliminaries of UDA-SS}
\label{sec:preliminary}
\vspace{-5pt}
We consider a source domain \textit{S} and a target domain \textit{T} in space \textit{x}$\times$\textit{y}, where \textit{x}, \textit{y} denote the input space and label space, respectively. In the source domain, we have $N_s$ images with labels ($x^S$=$\{x_i^s\}_{i=1}^{N_s}$, $y^S$=$\{y_i^s\}_{i=1}^{N_s}$) that belongs to \textit{C} categories, while for the target domain, we only have access to $N_t$ images $x^T$=$\{x_j^t\}_{j=1}^{N_t}$. A neural network consisting of a feature extractor \textit{$g_{\theta}$} and a segmentation head $h_{cls}$ is used as the adaptation model. The model is first trained on the source domain data \textit{$x^S$} with labels \textit{$y^S$}. Therefore segmentation loss function for the source domain becomes
\vspace{-5pt}
\begin{equation} \label{eq1}
L^{S}_{CE}=-\mathbb{E}[p^{s}_{i}\,\log\,h_{cls}\,(g_{\theta}(x^{s}_{i}))], 
\vspace{-5pt}
\end{equation}
where \textit{$p^{s}_{i}$} is the scalar value from one-hot vector of label \textit{$y^{s}_{i}$}. The value of \textit{$p^{s}_{i}(c)$} becomes 1 if, \textit{c} equals to \textit{$y^{s}_{i}$}, otherwise 0. 
Later a teacher network with the feature extractor $g_{\bar{\theta}}$ is used to predict $x^{t}_{j}$ and generate pseudo-label $\bar{y}^{t}_{j} = \textit{argmax} \allowbreak(h_{cls} \allowbreak g_{\bar{\theta}}(x^{t}_{j}))$.
 The teacher network is updated by the weight of the student model through {\em Exponential Moving Average (EMA)}. Though the target domain has no ground truth data, we can obtain a loss function $L^{T}_{CE}$ using the pseudo-labels. Mathematically,
\vspace{-2pt}
\begin{equation} \label{eq2}
L^{T}_{CE}=-\mathbb{E}[\bar{p}^{t}_{j}\,\log\,h_{cls}\,(g_{\bar{\theta}}(x^{t}_{j}))],
\vspace{-5pt}
\end{equation}
where \textit{$\bar{p}^{t}_{j}$} is the scalar value from one-hot vector of pseudo-label \textit{$\bar{y}^{t}_{j}$}. Based on Eq.~\eqref{eq1} and Eq.~\eqref{eq2}, the adaptation objective can be formulated as
\vspace{-2pt}
\begin{equation} \label{eq3}
\underset{\theta, \phi}{\min}L^{S}_{CE} (\theta, \phi)+L^{T}_{CE} (\theta, \phi),
\vspace{-5pt}
\end{equation}
where $\theta$, and $\phi$ are the weights of the feature extractor and segmentation head respectively.

\begin{figure}[t]
    \centering
    \includegraphics[width=0.9\linewidth]{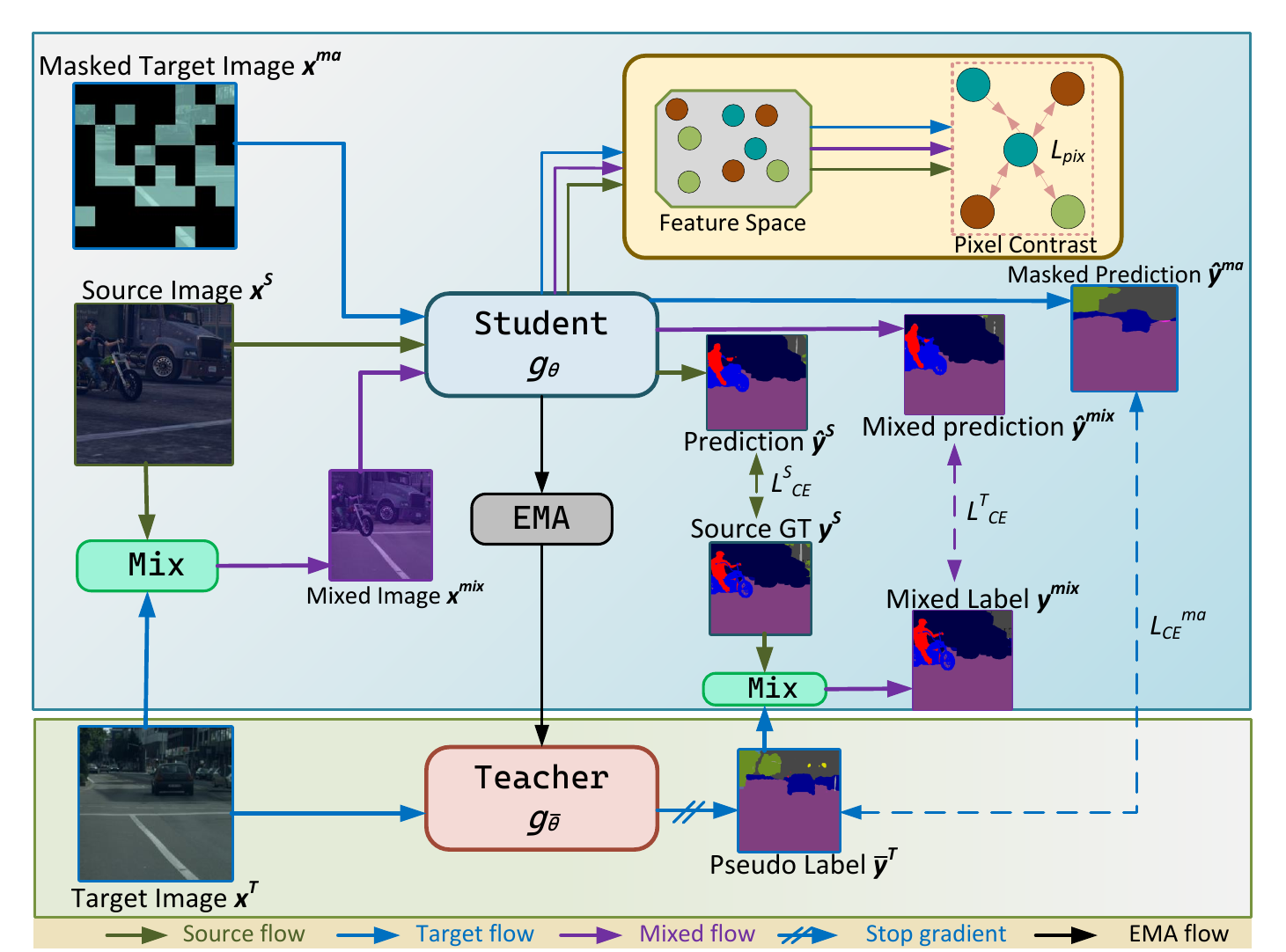} % Replace 'example-image' with your immixedage filename
    \caption{\small Framework overview of C$^2$DA. Given labeled source data \{$x^S$, $y^S$\}, we first calculate the source prediction $\hat{y}^{S}$ by using the student model. Later, we leverage the teacher model to predict pseudo-label $\bar{y}^{T}$. We craft the mixed label $y^{mix}$ and mixed data $x^{mix}$ by blending the images from both domains. We use the student model to predict mix prediction $\hat{y}^{mix}$. We also do the masking on target images to generate masked images $x^{ma}$ and leverage the student model to predict masked prediction images $\hat{y}^{ma}$ for learning contextual relations. Except for the segmentation losses we also use contrastive loss $L_{pix}$ for ensuring intra-class compactness and inter-class separability. }
    \vspace{-15pt}
    \label{pic1}
\end{figure}
\vspace{-5pt}
\subsection{Model Framework}
\label{sec:framework}
We build our proposed UDA framework by using the teacher-student framework (see Fig. \ref{pic1}). The architecture of the teacher model is the same as the student model, and both utilize a pre-trained SegFormer transformer. Each iteration consists of four stages. At the beginning of each iteration, we update the teacher model's weight by using an Exponential Moving Average (EMA), to ensure that the teacher model remains in sync with the student model. In the first stage, we train the student model with the source data and source label. After that, in the second stage, we use the teacher model to predict the target domain images and generate pseudo-labels without backpropagation. In the third stage, we use a cross-domain mix module (described in Sect. \ref{sec:context}) to generate a new image pair (\textit{$x^{mix}$}, \textit{$y^{mix}$}) from both domains and train the student model again. Hence, Eq. (\ref{eq2}) gets updated as:
\vspace{-10pt}
\begin{equation}
L^{T}_{CE}=-\mathbb{E}[p^{mix}_{j}\,\log\,h_{cls}\,(g_{\theta}(x^{mix}_{j}))],
\vspace{-5pt}
\end{equation}
where \textit{$p^{mix}_{j}$} is scalar value from the probability vector of the mixed label \textit{$y^{mix}_{j}$}, \textit{$p^{mix}_{j}$} is already one-hot encoded as we use copy-paste based mixing strategy. Finally, in the fourth stage, we mask out random patches from the target domain images (described in Sect. \ref{sec:masking}) and generate the masked images \textit{$x^{ma}$=$\{x_j^{ma}\}_{j=1}^{N_t}$}.
% For this, we generate a random mask \textit{M} from uniform distribution and perform element-wise multiplication between the \textit{M} and target image \textit{$x_{j}^{t}$}. Specifically, 
% \begin{equation}
% x_{j}^{ma}=M\otimes x^{t}_{j}. 
% \end{equation}
Then we train the student model again on the masked images supervised by the pseudo-labels \textit{$\bar{y}^{t}_{j}$}. %to predict the masked target prediction images $y^{ma}$. 
% In this way, the model can only access limited information from the unmasked regions of the target images, making the prediction more difficult. Hence, the model is forced to learn from the remaining context clues to reconstruct the pseudo-label \textit{$\bar{y}^{t}_{j}$} 
% \begin{equation}
% y^{ma}_{j}=h_{cls}(g_{\theta}(x_{j}^{ma})).
% \end{equation}
% As the pseudo-labels guide the model to generate masked target prediction, the masked loss \textit{$L^{ma}_{CE}$} can be formulated as:
% \begin{equation}
% L^{ma}_{CE}=-\mathbb{E}[\bar{p}^{t}_{j}\,log\,h_{cls}\,(g_{\theta}(x^{ma}_{j}))],
% \end{equation}
Based on the training of mixing and masking, the adaptation objective of our model is:
\vspace{-7pt}
\begin{equation} 
\underset{\theta, \phi}{\min}L^{S}_{CE} (\theta, \phi)+L^{T}_{CE} (\theta,\phi)+L^{ma}_{CE} (\theta, \phi),
\vspace{-5pt}
\end{equation}
where $L^{ma}_{CE}$ is the masked loss.
\vspace{-10pt}
\subsection{Learning Inherent Structure}
\label{sec:contrastive}
The adopted segmentation losses do not consider learning the inherent context within the images, which is important for local-focused segmentation tasks. So, to learn the intra-domain knowledge, we opt to utilize pixel-wise contrastive learning. Specifically, along with the classification head \textit{h$_{cls}$}, we use a projection head \textit{h$_{proj}$} that generates an embedding space \textit{es}=\textit{h$_{proj}$}$g_{\theta}(x)$ of the pixels. Contrastive learning facilitates learning the correlation between the labeled pixels by pulling the positive pairs of pixels together and pushing the negative pairs of pixels away. Considering the pixels of the same class \textit{C} as positive pairs and the other pixels in \textit{x} as negative pairs, contrastive loss \textit{$L_{pix}$} can be derived as
\vspace{-9pt}
\begin{equation} 
L_{pix}=-\underset{C(i)=C(j)}{\sum}\log\,\dfrac{d(es_{i},es_{j})}{\sum_{k=1}^{N_{pix}}d(es_{i},es_{k})},
\vspace{-5pt}
\end{equation}
where \textit{$N_{pix}$} is the total number of pixels, \textit{$es_{i}$} is the feature map of \textit{$i^{th}$} pixel in the embedding space, and \textit{d} denotes the similarity between the pixel features which can be measured using metrics like Euclidean distance or cosine similarity. In particular, we utilize the exponential form of cosine similarity 
\( d(es_{i}, es_{j}) = \exp(s(es_{i} , es_{j}) / \tau) \), where 
\( s \) represents the cosine similarity between two-pixel features 
\( es_{i} \) and 
\( es_{j} \), and 
\( \tau \) is the temperature parameter. The temperature parameter $\tau$ modulates the distribution sharpness of similarities. Effective implementation involves careful sampling of positive and negative pairs to ensure balanced training, efficient design of the embedding space to capture meaningful pixel relationships, and appropriate regularization techniques to prevent overfitting. As shown in Fig.~\ref{pic1}, after applying the contrastive loss, the feature space is guided to pull the pixels from the same class together and push the other away. This leads to a desirable property of intra-class compactness and inter-class separability ultimately enhancing segmentation performance by embedding richer contextual relationships.
\vspace{-15pt}
\subsection{Context-aware Mixing}\label{sec:context}
\vspace{-4pt}
Exploiting contextual knowledge is another important concept for mitigating the domain shift in UDA as both the source domain and target domain share similar semantic contexts. Domain shift refers to the discrepancy between the data distributions of the source and target domains, which can hinder model performance when applied to the target domain. By leveraging these shared semantic contexts, we create more meaningful training examples that help the model generalize better across domains. We use mixed images to calculate the target domain loss instead of the pseudo-labels as the mixture efficiently guides the model to learn from the supervision signals of both domains. According to ClassMix\cite{olsson2021classmix}, we first randomly choose and copy 50$\%$ of the classes from the source ground truth \textit{$y^{S}$} and then we paste them over the pseudo-label \textit{$\bar{y}^{T}$} to generate mixed label \textit{$y^{mix}$}. Similarly, we also paste the same class areas of \textit{$x^{S}$} over the \textit{$x^{T}$} to generate mixed image \textit{$x^{mix}$}. To be specific, we generate a mask, \textit{M}, by selecting random classes, and then apply this mask to the images from both domains, blending them to generate the mixed images. Formally,
\vspace{-7pt}
\begin{equation} 
\begin{split}
x^{mix}=M\otimes x^{S}+ (1-M) \otimes x^{T} \\
y^{mix}=M\otimes y^{S}+ (1-M) \otimes \bar{y}^{T}.
\end{split}
\vspace{-15pt}
\end{equation}
\vspace{-25pt}
\begin{figure}[h]
\centering
\subfloat[Source Image]{%
\includegraphics[width=0.25\textwidth]{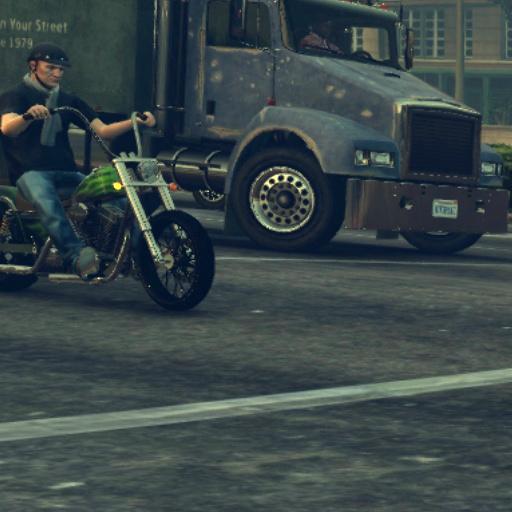}%
%\label{fig:middle}%
} 
\subfloat[Target Image]{%
\includegraphics[width=0.25\textwidth]{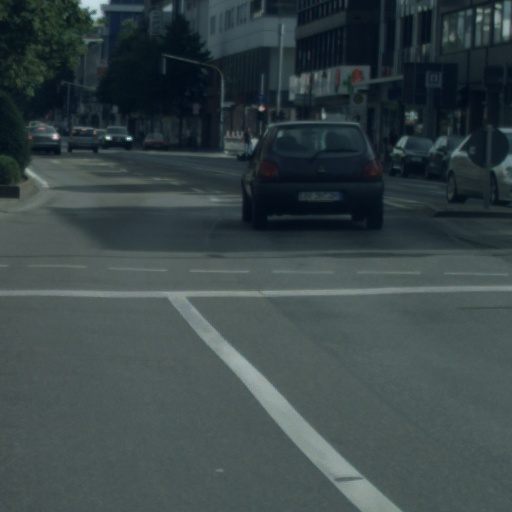}%
%\label{fig:left}%
} 
\subfloat[ClassMix]{%
\includegraphics[width=0.25\textwidth]{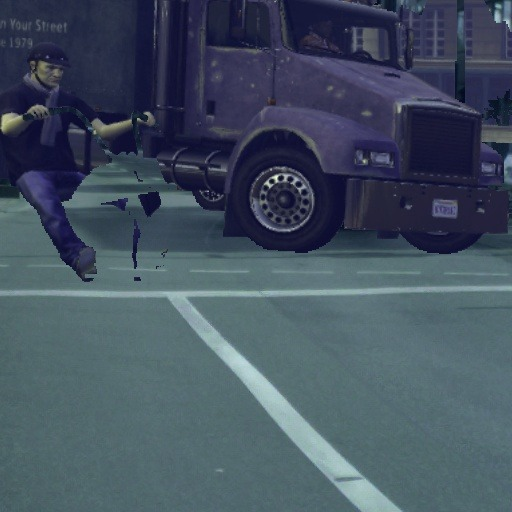}%
%\label{fig:left}%
}
\subfloat[Prior-guided Mix]{%
\includegraphics[width=0.25\textwidth]{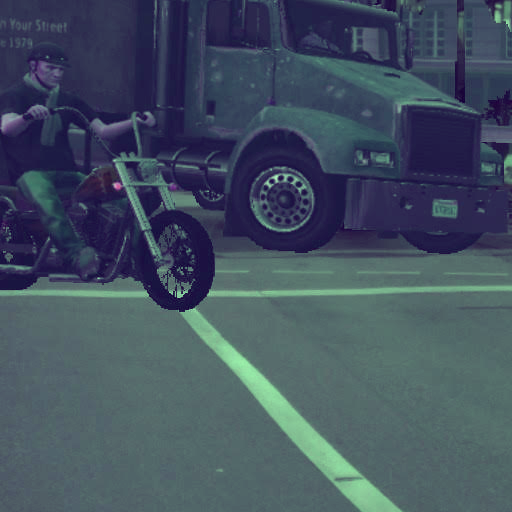}%
%\label{fig:middle}%
} \vspace{-10pt}
\caption{\small Illustration of the contextual advantage of the Prior-guided classmix over the conventional classmix.\vspace{-15pt}}\label{j} 
\end{figure}
However, in this way, the cross-domain mixture module overlooks the shared context across the domains. For example, in both the source and target domains, the \textit{rider} class is always associated with either \textit{bicycle} or \textit{motorcycle}. Moreover, the \textit{traffic light} is always beside the \textit{pole} and so on.  However, conventional mixing methods overlook this relationship due to random selection, copying the \textit{rider} class without including the \textit{bicycle} or \textit{motorcycle} class, thereby resulting in unrealistic mixing. There are a few other classes that also share similar semantic contexts according to the Cityscapes coarse annotation \cite{cordts2016cityscapes}. So, we modify the ClassMix in the name of \textit{Prior-Guided ClassMix} by using the coarse categories (e.g. object, vehicle) as the guidance in our work, to identify the classes with contextual relationships and copy these classes together from the source domain to paste at target domain images. Fig. \ref{j} demonstrates how the Prior-Guided ClassMix uses contextual relation to generate realistic mixed images.

Given the list of randomly chosen class list \textit{c} from the source ground truth \textit{$y^{S}$}, we check if each class \textit{K}$\in$\textit{c}, is also in the coarse category list \textit{l} or not. If it is then we append the semantically related classes \textit{$\bar{K}$} of the current class \textit{K} in the list \textit{c}. Then we use the updated \textit{c} to modify the mask \textit{M} and generate mixed images using this modified mask \textit{M}. By maintaining contextual relationships in the mixed images, our method creates more realistic and representative training examples. This helps the model learn features that are invariant to domain shift, thereby reducing the discrepancy between the source and target domains and improving the model's performance on the target domain.
\vspace{-10pt}
\subsection{Masking Module}
\label{sec:masking}
A masking module withholds local information from target images, encouraging the learning of context relations for robust recognition of classes with similar appearances by randomly masking out patches of target domain images. For masking out the random patches from target domain images, we generate a random mask $\mathcal{M}$  from a uniform distribution
\vspace{-7pt}
\begin{equation}
\mathcal{M}_{pa+1:(p+1)a, qb+1: (q+1)a} = [u > t] \quad \text{with} \quad u \sim \mathcal{R}(0, 1),
\vspace{-4pt}
\end{equation}
where the superscript of \( \mathcal{M} \) indicates the specific region of the image from rows \( pa+1 \) to \( (p+1)a \) and columns \( qb+1 \) to \( (q+1)b \). Here, \( a \) denotes the patch size, and \( t \) represents the mask ratio. The indices \( p \) and \( q \) range from \( 0 \) to \( \frac{W}{a} - 1 \), where \( W \) is the width of the image. The Height \( H \) is not shown explicitly here because the mask is applied in a symmetric manner across both the width and height of the image.
 Later we perform element-wise multiplication between the $\mathcal{M}$ and target image \textit{$x_{j}^{t}$}. Specifically, 
\vspace{-10pt}
\begin{equation}
x_{j}^{ma}=\mathcal{M}\otimes x^{t}_{j}. 
\vspace{-6pt}
\end{equation}
The student model is then retrained on masked images using pseudo-labels \textit{$\bar{y}^{t}_{j}$} to predict masked target images $y^{ma}$. In this way, the model can only access limited information from the unmasked regions of the target images, making the prediction more difficult. Hence, the model is forced to learn from the remaining context clues to reconstruct the pseudo-label \textit{$\bar{y}^{t}_{j}$} 
\vspace{-8pt}
\begin{equation}
y^{ma}_{j}=h_{cls}(g_{\theta}(x_{j}^{ma})).
\vspace{-5pt}
\end{equation}
As the pseudo-labels guide the model to generate masked target prediction, the masked loss \textit{$L^{ma}_{CE}$} can be formulated as:
\vspace{-9pt}
\begin{equation}
L^{ma}_{CE}=-\mathbb{E}[\bar{p}^{t}_{j}\,\log\,h_{cls}\,(g_{\theta}(x^{ma}_{j}))]. 
\vspace{-10pt}
\end{equation}
\vspace{-10pt}
\section{Experiments}
\vspace{-5pt}
\subsection{Evaluation Setup}
\normalsize \textbf{Datasets:} We have tested our model in five datasets. (1)\textbf{ GTA-V} is a simulation dataset collected in the city environment. The dataset consists of 24,966 synthetic images with a resolution of 1914$\times$1052. The dataset has annotated labels for 33 classes. (2) \textbf{Synthia} dataset is another city-like simulation dataset. This dataset consists of 9400 images with a resolution of 1280$\times$760. (3) \textbf{Cityscapes} dataset has 2975 training images and 500 validation images. All the images have been collected from European cities. (4) \textbf{RUGD} is an off-road dataset to improve robot navigation in unstructured environments. The dataset has 7453 labeled images containing 24 semantic categories and eight unique terrain types. (5) \textbf{MESH} dataset is another forest dataset consisting of 2612 training and 58 validation images. We have performed two sim2real applications\textemdash from GTA V$\rightarrow$cityscapes dataset, Synthia$\rightarrow$cityscapes dataset, and one forset adaptation from RUGD$\rightarrow$MESH dataset. We have employed mean Intersection over Union (mIoU) as the performance metric for city adaptations and have utilized qualitative results to gauge the performance of forest adaptations, as the MESH dataset lacks available ground truths. Adaptations have been evaluated in the validation sets of Cityscapes and MESH datasets.
\newline \textbf{Implementation Details:}
We have based our UDA method on the self-training framework in recent SOTA work MIC \cite{hoyer2023mic}. We used a batch size of 1 and set a crop size of 952 due to limited GPU memory. 
%We have compared our model with the MIC and other SOTA methods. To perform a valid comparison we also run the baseline in our setup.
\vspace{-15pt}
\subsection{Comparison}
\begin{figure}[t] 
{
    \centering
    \includegraphics[width=1\linewidth]{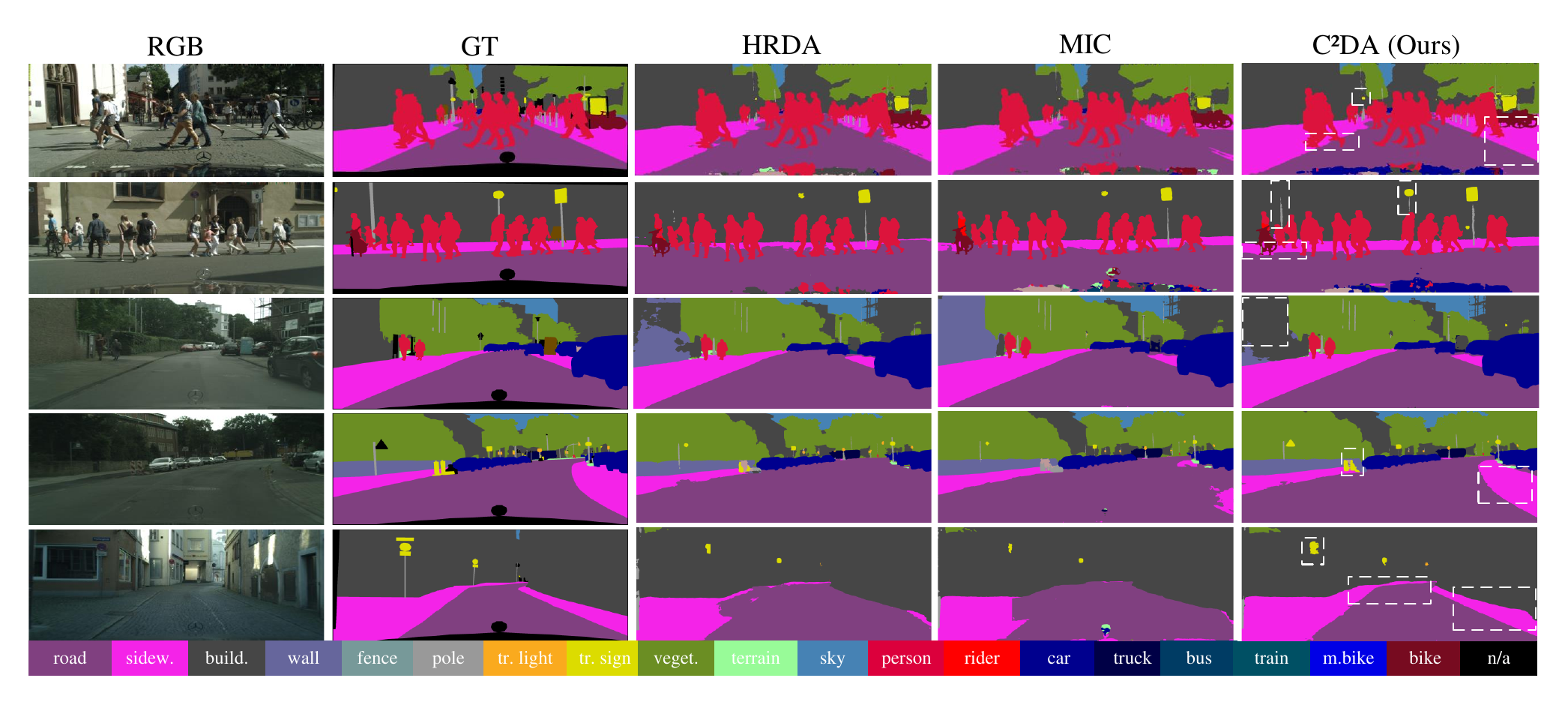} \vspace{-25pt}% Replace 'example-image' with your image filename
    \caption{\small Qualitative Results for the adaptation of GTA-V → Cityscapes. \vspace{-20pt}}
    \label{pic2}
}
\end{figure}

\begin{figure}[h]
{
    \centering
    \includegraphics[width=1\linewidth]{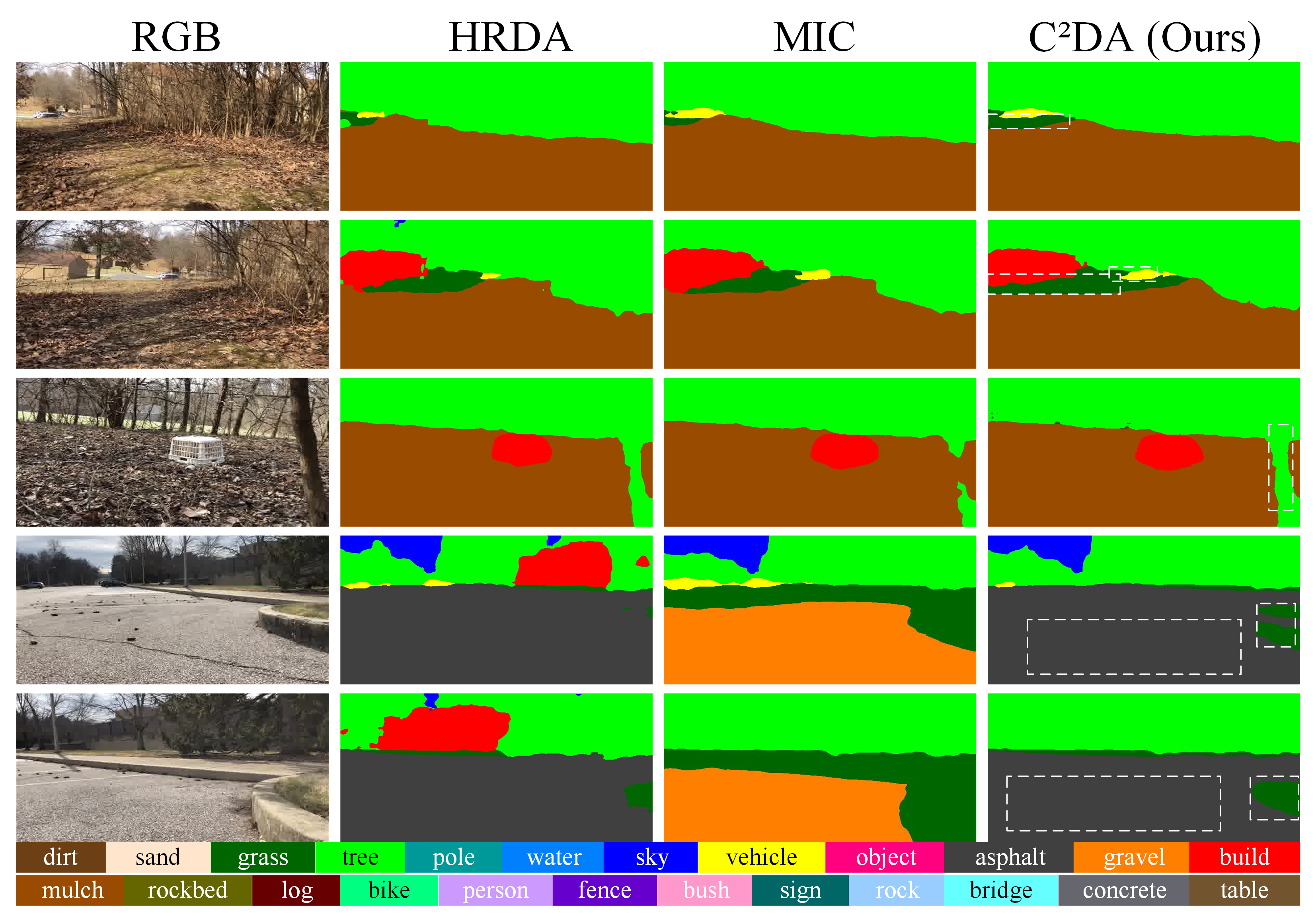}  \vspace{-25pt}
    \caption{Qualitative Results for the adaptation of RUGD → MESH.  \vspace{-20pt}}
    \label{pic3}
}
\end{figure}

\vspace{-5pt}
We compare our proposed UDA framework with the baseline method MIC in both quantitative and qualitative manner. Besides the baseline, we perform a quantitative comparison with other SOTA methods as well. First, we show the quantitative comparison of GTA V$\rightarrow$Cityscapes adaptation in Table \ref{tab1}.
\vspace{-10pt}
\begin{table}[h]
\scriptsize
    \centering
    \addtolength{\tabcolsep}{0pt}
    \vspace{-15pt}
    \caption{\small Quantitative comparison for the adaptation of GTA-V $\rightarrow$ Cityscapes with SOTA methods including FDA\cite{yang2020fda}, PIT\cite{lv2020cross}, IAST\cite{mei2020instance}, DACS\cite{tranheden2021dacs}, CorDA\cite{wang2021domain}, ProDA\cite{zhang2021prototypical}, IDA\cite{chen2023ida}, DAFormer\cite{hoyer2022daformer}, HRDA\cite{hoyer2022hrda}, MIC\cite{hoyer2023mic}. The \textcolor{teal!70}{teal} and \textcolor{purple!70}{purple} show the best and second-best results respectively.}
    \label{tab1}
    \begin{tabular}{l||ccccccccccc}
        \toprule
        % Class & FDA\cite{yang2020fda} & PIT\cite{lv2020cross} & IAST\cite{mei2020instance} & DACS\cite{tranheden2021dacs} & CorDA\cite{wang2021domain} & ProDA\cite{zhang2021prototypical} & IDA\cite{chen2023ida} & DAFormer\cite{hoyer2022daformer} & HRDA\cite{hoyer2022hrda} & MIC\cite{hoyer2023mic} & C$^2$DA (Ours) \\
        Class & FDA & PIT & IAST & DACS& CorDA & ProDA & IDA & DAFormer & HRDA & MIC & C$^2$DA (Ours) \\ \hline
        \midrule
        Road & 92.5 & 87.5 & 93.8 & 89.9 & 94.7 & 87.8 & 95.4 & 95.7 & 96.59 & \textbf{\cellcolor{purple!20}97.13} & \textbf{\cellcolor{teal!20}97.57} \\ 
        S.Walk & 53.3 & 43.4 & 57.8 & 39.7 & 63.1 & 56.0 & 72.0 & 70.2 & 74.32 & \textbf{\cellcolor{purple!20}77.8} & \textbf{\cellcolor{teal!20}79.5} \\ 
        Build & 82.4 & 78.8 & 85.1 & 87.9 & 87.6 & 79.7 & 87.8 & 89.4 & 89.06 & \textbf{\cellcolor{purple!20}90.17} & \textbf{\cellcolor{teal!20}90.69} \\ 
        Wall & 26.5 & 31.2 & 39.5 & 30.7 & 30.7 & 46.3 & 49.9 & 53.5 & 56.55 & \textbf{\cellcolor{purple!20}57.27} & \textbf{\cellcolor{teal!20}57.69} \\ 
        Fence & 27.6 & 30.2 & 26.7 & 39.5 & 40.6 & 44.8 & 36.6 & 48.1 & 39.61 & \textbf{\cellcolor{teal!20}53.58} & \textbf{\cellcolor{purple!20}52.91} \\ 
        Pole & 36.4 & 36.3 & 26.2 & 38.5 & 40.2 & 45.6 & 40.6 & 49.6 & 50.92 & \textbf{\cellcolor{purple!20}51.48} & \textbf{\cellcolor{teal!20}53.99} \\ 
        T. Light & 40.6 & 39.9 & 43.1 & 46.4 & 47.8 & 53.5 & 46.8 & 55.8 & \textbf{\cellcolor{teal!20}59.05} & \textbf{\cellcolor{purple!20}58.51} & 58.31 \\ %\hline
        Sign & 38.9 & 42.0 & 34.7 & 52.8 & 51.6 & 53.5 & 50.4 & \textbf{\cellcolor{purple!20}59.4} & 58.1 & 50.45 & \textbf{\cellcolor{teal!20}65.17} \\ %\hline
        Veg & 82.3 & 79.2 & 84.9 & 88.0 & 87.6 & 88.6 & 88.3 & 89.9 & 90.41 & \textbf{\cellcolor{purple!20}90.51} & \textbf{\cellcolor{teal!20}91.03} \\ %\hline
        Terrian & 39.8 & 37.1 & 32.9 & 44.0 & 47.0 & 45.2 & 45.2 & 47.9 & \textbf{\cellcolor{teal!20}49.7} & 49.22 & \textbf{\cellcolor{purple!20}49.35} \\ %\hline
        Sky & 78.0 & 79.3 & 88.0 & 88.8 & 89.7 & 82.1 & 92.1 & 92.5 & \textbf{\cellcolor{purple!20}94.12} & \textbf{\cellcolor{teal!20}94.68} & 93.87 \\ %%\hline
        Person & 62.6 & 65.4 & 62.6 & 67.2 & 66.7 & 70.7 & 74.2 & 72.2 & 76.3 & \textbf{\cellcolor{purple!20}76.6} & \textbf{\cellcolor{teal!20}77.78} \\ %\hline
        Rider & 34.4 & 37.5 & 29.0 & 35.8 & 35.9 & 39.2 & 50.4 & 44.7 & 49.39 & \textbf{\cellcolor{teal!20}52.17} & \textbf{\cellcolor{purple!20}51.76} \\ %\hline
        Car & 84.9 & 83.2 & 87.3 & 84.5 & 90.2 & 88.8 & 92.8 & 92.3 & 93.43 & \textbf{\cellcolor{purple!20}93.85} & \textbf{\cellcolor{teal!20}93.99} \\ %\hline
        Truck & 34.1 & 46.0 & 39.2 & 45.7 & 48.9 & 45.5 & 79.2 & 74.5 & \textbf{\cellcolor{purple!20}82.45} & \textbf{\cellcolor{teal!20}82.56} & 81.28 \\ %\hline
        Bus & 53.1 & 45.6 & 49.6 & 50.2 & 57.5 & 59.4 & 81.8 & 78.2 & 68.48 & \textbf{\cellcolor{teal!20}86.76} & \textbf{\cellcolor{purple!20}85.95} \\ %\hline
        Train & 16.9 & 25.7 & 23.2 & 0.0 & 0.0 & 1.0 & 53.8 & 65.1 & 1.87 & \textbf{\cellcolor{teal!20}76.83} & \textbf{\cellcolor{purple!20}67.99} \\ %\hline
        MC & 27.7 & 23.5 & 34.7 & 27.3 & 39.8 & 48.9 & 61.4 & 55.9 & 61.94 & \textbf{\cellcolor{teal!20}62.65} & \textbf{\cellcolor{purple!20}61.94} \\ %\hline
        Bike & 46.4 & 49.9 & 39.6 & 34.0 & 56.0 & 56.4 & 64.5 & 61.8 & 66.95 & \textbf{\cellcolor{purple!20}67.23} & \textbf{\cellcolor{teal!20}68.5} \\ %\hline
        mIoU & 50.45 & 50.6 & 51.5 & 52.1 & 56.6 & 57.5 & 66.5 & 68.3 & 66.3 & \textbf{\cellcolor{purple!20}72.08} & \textbf{\cellcolor{teal!20}72.59} \\ 
        \bottomrule
    \end{tabular}
    \vspace{-15pt}
\end{table}
\normalsize The table shows that our model outperforms all the SOTA methods in terms of mIoU. 
Our method also wins over the baseline by a margin of 0.51$\%$ mIoU. Our model shows a superior performance in 10 out of 19 classes including challenging classes like \textit{Wall}, \textit{Person}, \textit{Traffic sign}, \textit{Vegetation}, \textit{Bike} etc. The comparison result of Synthia$\rightarrow$Cityscapes adaptation is presented in Table \ref{tab01}. 
%\vspace{-5pt}
\vspace{-25pt}
\begin{table}[h]
    \centering
   
    \caption{\small Quantitative comparison for the adaptation of Synthia $\rightarrow$ Cityscapes with SOTA methods including ADVENT\cite{vu2019advent}, PIT\cite{lv2020cross}, IAST\cite{mei2020instance}, DACS\cite{tranheden2021dacs}, ProDA\cite{zhang2021prototypical}, IDA\cite{chen2023ida}, DAFormer\cite{hoyer2022daformer}, HRDA\cite{hoyer2022hrda}, MIC\cite{hoyer2023mic}. The \textcolor{teal!70}{teal} and \textcolor{purple!70}{purple} colors show the best and second-best results respectively.}
    \scriptsize
    
    \label{tab01}
    \setlength{\tabcolsep}{1pt}
    \begin{tabular}{l||cccccccccc}
        \toprule
        % Class & ADVENT\cite{vu2019advent} & PIT\cite{lv2020cross} & IAST\cite{mei2020instance} & DACS\cite{tranheden2021dacs} & ProDA\cite{zhang2021prototypical} & IDA\cite{chen2023ida} & DAFormer\cite{hoyer2022daformer} & HRDA\cite{hoyer2022hrda} & MIC\cite{hoyer2023mic} & C$^2$DA (Ours) \\

        Class & ADVENT & PIT & IAST& DACS & ProDA& IDA & DAFormer& HRDA & MIC & C$^2$DA (Ours) \\
        
        \midrule
        Road & 85.6 & 83.1 & 81.9 & 80.5 & 87.8 & 88.9 & 84.5 & 87.25 & \textbf{\cellcolor{purple!20}95.9} & \textbf{\cellcolor{teal!20}96.31} \\ %\hline
        S.Walk & 42.2 & 27.6 & 41.5 & 25.1 & 45.7 & 44.2 & 40.7 & 42.02 & \textbf{\cellcolor{purple!20}71.06} & \textbf{\cellcolor{teal!20}75.22} \\ %\hline
        Build & 79.7 & 81.5 & 83.3 & 81.9 & 84.6 & 78.2 & \textbf{\cellcolor{purple!20}88.4} & \textbf{\cellcolor{teal!20}88.88} & 88.05 & 88.38 \\ %\hline
        Wall & 8.7 & 8.9 & 17.7 & 21.4 & 37.1 & \textbf{\cellcolor{teal!20}49.1} & 41.5 & \textbf{\cellcolor{purple!20}46.33} & 35.13 & 41.42 \\ %\hline
        Fence & 0.4 & 0.3 & 4.6 & 2.8 & 0.6 & 4.9 & \textbf{\cellcolor{purple!20}6.5} & 2.33 & 3.79 & \textbf{\cellcolor{teal!20}7.71} \\ %\hline
        Pole & 25.9 & 21.8 & 32.3 & 37.2 & 44.0 & 48.6 & 50.0 & \textbf{\cellcolor{purple!20}52.26} & \textbf{\cellcolor{teal!20}53.12} & 51.92 \\ %\hline
        T. Light & 5.4 & 26.4 & 30.9 & 22.6 & 54.6 & 52.3 & 55.0 & \textbf{\cellcolor{purple!20}60.94} & \textbf{\cellcolor{teal!20}61.04} & 60.9 \\ %\hline
        Sign & 8.1 & 33.8 & 28.8 & 23.9 & 37.0 & 49.3 & \textbf{\cellcolor{purple!20}54.6} & 51.35 & 54.13 & \textbf{\cellcolor{teal!20}61.77} \\ %\hline
        Veg & 80.4 & 76.4 & 83.4 & 83.6 & 88.1 & 84.9 & 86.0 & 88.45 & \textbf{\cellcolor{teal!20}89.75} & \textbf{\cellcolor{purple!20}88.93} \\ %\hline
        Sky & 84.1 & 78.8 & 85.0 & 90.7 & 84.4 & 88.2 & 89.8 & 94.05 & \textbf{\cellcolor{purple!20}94.56} & \textbf{\cellcolor{teal!20}94.64} \\ %\hline
        Person & 57.9 & 64.2 & 65.5 & 67.6 & 74.2 & 70.1 & 73.2 & 78.92 & \textbf{\cellcolor{purple!20}79.2} & \textbf{\cellcolor{teal!20}79.91} \\ %\hline
        Rider & 23.8 & 27.6 & 30.8 & 38.3 & 24.3 & 47.0 & 48.2 & 52.52 & \textbf{\cellcolor{teal!20}54.98} & \textbf{\cellcolor{purple!20}53.49} \\ %\hline
        Car & 73.3 & 79.6 & 86.5 & 82.9 & 88.2 & 85.3 & 87.2 & 89.39 & \textbf{\cellcolor{purple!20}88.97} & \textbf{\cellcolor{teal!20}89.47} \\ %\hline
        Bus & 36.4 & 31.2 & 38.2 & 38.9 & 51.1 & 58.2 & 53.2 & \textbf{\cellcolor{purple!20}60.87} & \textbf{\cellcolor{teal!20}60.93} & 55.04 \\ %\hline
        MC & 14.2 & 31.0 & 33.1 & 28.4 & 40.5 & 58.7 & 53.9 & 60.39 & \textbf{\cellcolor{teal!20}65.93} & \textbf{\cellcolor{purple!20}62.34} \\ %\hline
        Bike & 33.0 & 31.3 & 52.7 & 47.5 & 45.6 & 56.9 & 61.7 & \textbf{\cellcolor{teal!20}64.16} & \textbf{\cellcolor{purple!20}62.38} & 60.16 \\ %\hline
        mIoU & 41.2 & 44.0 & 49.8 & 48.3 & 55.5 & 60.3 & 60.9 & 63.76 & \textbf{\cellcolor{purple!20}66.18} & \textbf{\cellcolor{teal!20}66.72} \\ 
        \bottomrule
    \end{tabular}
    \vspace{-20pt}
\end{table}
\normalsize From this table, we can see that our model outperforms the strong baseline MIC with a margin of 0.54$\%$ mIoU. Our model also reveals a better performance in challenging classes like \textit{Fence}, \textit{Traffic sign}, \textit{Person} etc. It is worth noting that the results of HRDA and MIC shown in Table \ref{tab1} and Table \ref{tab01} are lower than the one reported in the original works, due to adjustments in hyperparameter settings to conduct a fair comparison with our work. The superiority of our proposed UDA framework is further demonstrated through the examples presented in Fig. \ref{pic2} and Fig. \ref{pic3} for GTA V$\rightarrow$Cityscapes adaptation and RUGD$\rightarrow$MESH adaptation, respectively.
\vspace{-10pt}
\begin{table} [h]
\footnotesize
  \centering
  %\vspace{-5pt}
  \caption{\small Comparison of different learning components. PG stands for \textit{Prior-Guided}. }
  \label{tab3}
  \setlength{\tabcolsep}{1.5pt}
  \begin{tabular}{l|ccc}
    \hline
    Component & mIoU & $\delta_{MIC}$ & $\delta_{OUR}$ \\
    \hline
    \hline
    PG ClassMix+Contrastive learning & 69.08 & -3 & -3.51 \\
    Masking+Contrastive learning & 70.54 & -1.54 & -2.05 \\
    Masking+PG ClassMix & 72.41 & 0.33 & -0.18 \\
    Masking+PG ClassMix+Contrastive learning & 72.59 & 0.51 & 0 \\
    \hline 
  \end{tabular}
  \vspace{-10pt}
\end{table}
\vspace{-10pt}
\subsection{Ablation Studies}
\vspace{-5pt}
\subsubsection{Effect of Each Module}
\normalsize In the proposed UDA framework, we adopt three different modules, i.e., a Prior-Guided ClassMix module, a contrastive learning module, and a masking module. We evaluate the necessity of each of these modules by ablating that module. Our complete framework achieves 72.59 mIoU, which is 0.51$\%$ higher than the baseline MIC. However, the performance of our framework drops to 69.08 (see Table \ref{tab3}), if we remove the masking module which demonstrates the heavy influence of learning from context relations using masked images. Replacing the Prior-Guided Classmix technique with the original ClassMix in our work, our method obtains a mIoU of 70.54, which is still far from the result of our complete framework, implying that generating realistic mixed images helps our model to perform better. We also evaluate the effectiveness of the contrastive learning strategy by removing it from our framework. We observe our model achieves 72.41 mIoU without contrastive learning. From the result, we can conclude that the effect of contrastive learning is not as high as the masking module and the Prior-Guided ClassMix, however still this module helps to increase the result by a mIoU of 0.18.
\vspace{-12pt}
\subsubsection{Prior-Guided ClassMix Variations}
In our framework, we modify the ClassMix technique by utilizing the prior knowledge from the coarse categories of the Cityscapes dataset. However, we do not consider all the coarse groups together in one experiment as the excessive use of prior knowledge can hamper the performance of our proposed UDA framework. We run several experiments with different combinations of the coarse categories. Table \ref{tab4} shows the influence of the different combinations of coarse categories. Compared to our baseline MIC, we only obtain lower results when we use the combination of Flat (\textit{road}, \textit{sidewalk}) and Nature (\textit{vegetation}, \textit{terrain}) coarse categories. The reason behind this performance drop can be the discrepancy between the spatial relations of these two coarse categories.
The best performance is achieved for the combination of Construction (\textit{building}, \textit{wall}, \textit{fence}) and Nature coarse categories. From our observation, this combination is giving us the best result, because along with the intra-coarse category relation, it has an inter-coarse category relation as well. In particular, \textit{building} is always near the \textit{vegetation} and the \textit{fence} or \textit{wall} is always near the \textit{terrain}. So, this combination produces the most realistic mixing of the images. 
\vspace{-20pt}
\begin{table}
    \centering

    \caption{Comparison of Different Combinations of the Coarse Categories. } 
    \label{tab4}
   \begin{tabular}{l|ccc}
    \hline
    Combinations & mIoU & $\delta_{MIC}$ & $\delta_{OUR}$ \\
         \hline
    \hline
        
       Flat, Nature & 71.46 & -0.62& -1.13 \\
       
       Objects, Human-Vehicle&72.43&0.35&-0.16\\
      
       Construction, Nature&72.59&0.51&0\\
       \hline
       
\end{tabular}
\vspace{-20pt}
\end{table}
\vspace{-15pt}
\subsubsection{Contrastive learning Variations}
In the UDA framework, we adapt contrastive learning to learn the inherent structures of the images. In the proposed work, we utilize the student model for supervised learning in three types of images, e.g., source domain images, mixed images, and masked target images. So, we have applied contrastive loss in all these three stages. However, to show the influence of learning intra-domain knowledge we perform several experiments by applying the contrastive loss in different ways. Table \ref{tab5} shows the quantitative results for the usage of contrastive loss in our work. From Table \ref{tab5}, it can be seen that the result gradually increases if we consider multiple stages for learning the intra-domain knowledge. However, the combination of Source domain+Mixing is better than the combination of Source domain+Masking.
\vspace{-30pt}
\begin{table}[h]
\centering
\caption{Comparison of Different Combinations of the Contrastive Loss. } 
\label{tab5}
\begin{tabular}{l|ccc}
\hline
Combinations & mIoU &$\delta_{MIC}$ & $\delta_{OUR}$ \\
   \hline
   \hline
   Source domain & 69.58&-2.5&-3.01  \\
   
   Source Domain+ Masking&70.43&-1.65&-2.16\\
 
   Source domain+Mixing&71.4&-0.68&-1.19\\
   
   Source domain+Mixing+ Masking&72.59&0.51&0\\
   \hline
\end{tabular}
\vspace{-15pt}
\end{table}
%\vspace{14pt}
\begin{figure} [h]
{
    \centering
    \includegraphics[width=1\linewidth]{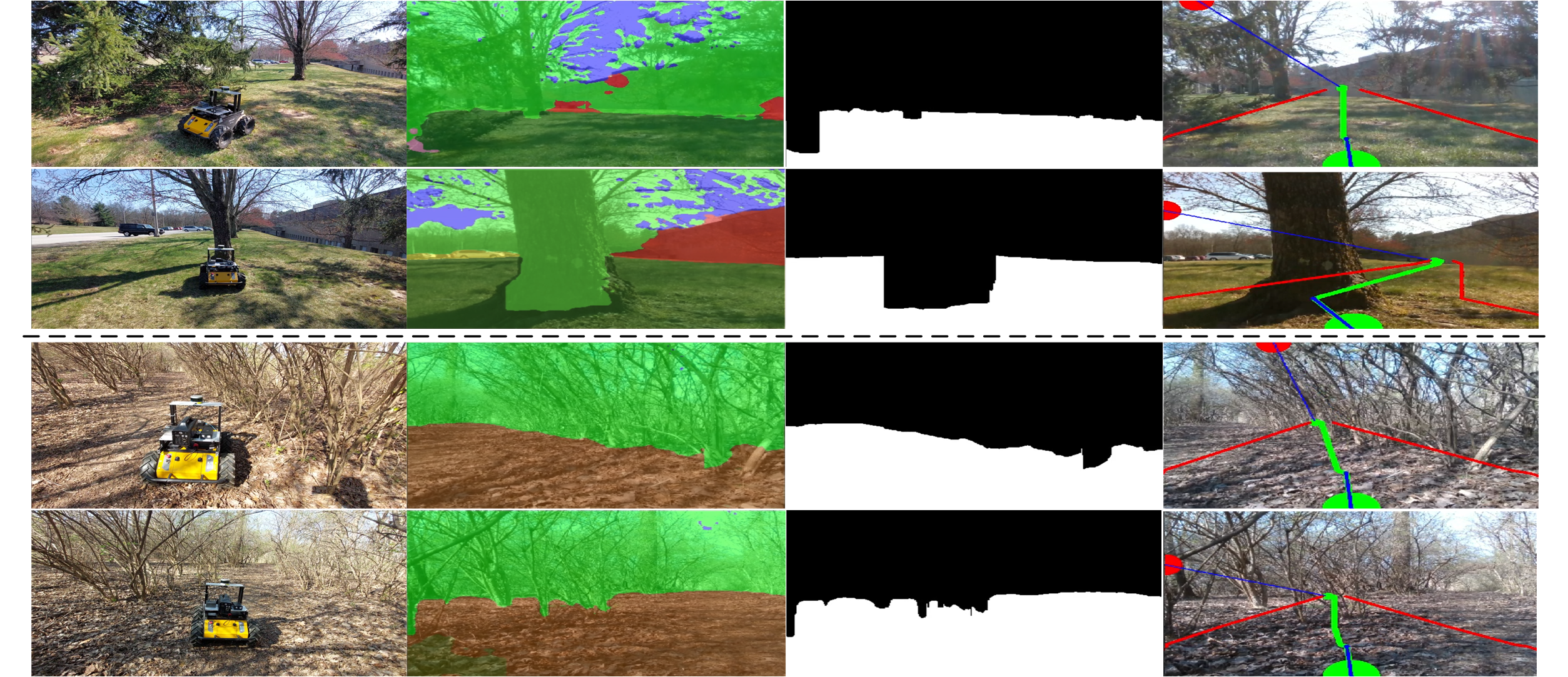} \vspace{-10pt}
    \caption{Navigation behavior in the forest environment. The first two rows show the scenario 1 and the last two rows show the scenario 2. We show the third-person view, segmentation result, navigable image result, and the planning result from the left column to the right column respectively. \label{pic4}  \vspace{-18pt}}   
}
\end{figure}
\vspace{-10pt}
\subsection{Navigation Missions}
We combine our proposed C$^2$DA model (trained with RUGD$\rightarrow$MESH setup) with POVNav planner \cite{pushp2023povnav} to show the effectiveness of our model in real-world deployment. We examine the behavior of our navigation system in two different forest scenarios (see Fig. \ref{pic4}), where the first scenario consists of grass and trees and the second scenario consists of mulch and trees. 
%\vspace{-10pt}

During the navigation, the image resolution is set to 640$\times$480 pixels. We deploy the navigation stack on a Husky robot with an NVIDIA GeForce RTX 2060 GPU. Our proposed C$^2$DA takes 0.54 seconds to perform pure segmentation. As a complete perception system requires some other processing rather than segmentation, our whole process takes around 0.60 seconds. We set the linear velocity of the vehicle as 0.3\textit{m/s} and set the path length for the two scenarios as 15m and 9m respectively. Though the navigation speed was slow, our method shows expected behavior by avoiding the non-navigable classes and reaching the goal. This demonstrates that our proposed C$^2$DA model is capable of performing navigation tasks in unstructured environments. 
\vspace{-10pt}
\section{CONCLUSIONS}
\vspace{-7pt}
We present a model named C$^2$DA to improve unsupervised domain adaptive semantic segmentation. Our approach is a self-training framework that explores both the inherent structures of the images and the contextual clues of the target domain images. To achieve this, we incorporate contrastive loss along with traditional cross-entropy loss. Furthermore, we modify the ClassMix technique by leveraging the contextual dependency of the classes and applying the masking technique to the target domain images to ensure context-aware learning. Evaluation of benchmark datasets demonstrates that our model outperforms the SOTA by a slight margin. Moreover, we deployed our framework in a robotic vehicle to navigate in unstructured environments.
{\small 
\vspace{-5pt}
\bibliographystyle{IEEEtran}
\bibliography{ref}

\begin{thebibliography}{10}
\providecommand{\url}[1]{#1}
\csname url@samestyle\endcsname
\providecommand{\newblock}{\relax}
\providecommand{\bibinfo}[2]{#2}
\providecommand{\BIBentrySTDinterwordspacing}{\spaceskip=0pt\relax}
\providecommand{\BIBentryALTinterwordstretchfactor}{4}
\providecommand{\BIBentryALTinterwordspacing}{\spaceskip=\fontdimen2\font plus
\BIBentryALTinterwordstretchfactor\fontdimen3\font minus \fontdimen4\font\relax}
\providecommand{\BIBforeignlanguage}[2]{{%
\expandafter\ifx\csname l@#1\endcsname\relax
\typeout{** WARNING: IEEEtran.bst: No hyphenation pattern has been}%
\typeout{** loaded for the language `#1'. Using the pattern for}%
\typeout{** the default language instead.}%
\else
\language=\csname l@#1\endcsname
\fi
#2}}
\providecommand{\BIBdecl}{\relax}
\BIBdecl

\bibitem{cordts2016cityscapes}
M.~Cordts, M.~Omran, S.~Ramos, T.~Rehfeld, M.~Enzweiler, R.~Benenson, U.~Franke, S.~Roth, and B.~Schiele, ``The cityscapes dataset for semantic urban scene understanding,'' in \emph{Proceedings of the IEEE conference on computer vision and pattern recognition}, 2016, pp. 3213--3223.

\bibitem{sakaridis2021acdc}
C.~Sakaridis, D.~Dai, and L.~Van~Gool, ``Acdc: The adverse conditions dataset with correspondences for semantic driving scene understanding,'' in \emph{IEEE/CVF International Conference on Computer Vision}, 2021, pp. 10\,765--10\,775.

\bibitem{ros2016synthia}
G.~Ros, L.~Sellart, J.~Materzynska, D.~Vazquez, and A.~M. Lopez, ``The synthia dataset: A large collection of synthetic images for semantic segmentation of urban scenes,'' in \emph{Proceedings of the IEEE conference on computer vision and pattern recognition}, 2016, pp. 3234--3243.

\bibitem{gong2021dlow}
R.~Gong, W.~Li, Y.~Chen, D.~Dai, and L.~Van~Gool, ``Dlow: Domain flow and applications,'' \emph{International Journal of Computer Vision}, vol. 129, no.~10, pp. 2865--2888, 2021.

\bibitem{hoyer2022daformer}
L.~Hoyer, D.~Dai, and L.~Van~Gool, ``Daformer: Improving network architectures and training strategies for domain-adaptive semantic segmentation,'' in \emph{Proceedings of the IEEE/CVF Conference on Computer Vision and Pattern Recognition}, 2022, pp. 9924--9935.

\bibitem{sajjadi2016regularization}
M.~Sajjadi, M.~Javanmardi, and T.~Tasdizen, ``Regularization with stochastic transformations and perturbations for deep semi-supervised learning,'' \emph{Advances in neural information processing systems}, vol.~29, 2016.

\bibitem{chen2023pipa}
M.~Chen, Z.~Zheng, Y.~Yang, and T.-S. Chua, ``Pipa: Pixel-and patch-wise self-supervised learning for domain adaptative semantic segmentation,'' in \emph{Proceedings of the 31st ACM International Conference on Multimedia}, 2023, pp. 1905--1914.

\bibitem{hoyer2023mic}
L.~Hoyer, D.~Dai, H.~Wang, and L.~Van~Gool, ``Mic: Masked image consistency for context-enhanced domain adaptation,'' in \emph{Proceedings of the IEEE/CVF Conference on Computer Vision and Pattern Recognition}, 2023, pp. 11\,721--11\,732.

\bibitem{olsson2021classmix}
V.~Olsson, W.~Tranheden, J.~Pinto, and L.~Svensson, ``Classmix: Segmentation-based data augmentation for semi-supervised learning,'' in \emph{Proceedings of the IEEE/CVF Winter Conference on Applications of Computer Vision}, 2021, pp. 1369--1378.

\bibitem{ganin2016domain}
Y.~Ganin, E.~Ustinova, H.~Ajakan, P.~Germain, H.~Larochelle, F.~Laviolette, M.~March, and V.~Lempitsky, ``Domain-adversarial training of neural networks,'' \emph{Journal of machine learning research}, vol.~17, no.~59, pp. 1--35, 2016.

\bibitem{chen2021scale}
Y.~Chen, H.~Wang, W.~Li, C.~Sakaridis, D.~Dai, and L.~Van~Gool, ``Scale-aware domain adaptive faster r-cnn,'' \emph{International Journal of Computer Vision}, vol. 129, no.~7, pp. 2223--2243, 2021.

\bibitem{hoffman2018cycada}
J.~Hoffman, E.~Tzeng, T.~Park, J.-Y. Zhu, P.~Isola, K.~Saenko, A.~Efros, and T.~Darrell, ``Cycada: Cycle-consistent adversarial domain adaptation,'' in \emph{International conference on machine learning}.\hskip 1em plus 0.5em minus 0.4em\relax Pmlr, 2018, pp. 1989--1998.

\bibitem{hoyer2022hrda}
L.~Hoyer, D.~Dai, and L.~Van~Gool, ``Hrda: Context-aware high-resolution domain-adaptive semantic segmentation,'' in \emph{European Conference on Computer Vision}.\hskip 1em plus 0.5em minus 0.4em\relax Springer, 2022, pp. 372--391.

\bibitem{long2016unsupervised}
M.~Long, H.~Zhu, J.~Wang, and M.~I. Jordan, ``Unsupervised domain adaptation with residual transfer networks,'' \emph{Advances in neural information processing systems}, vol.~29, 2016.

\bibitem{sun2016deep}
B.~Sun and K.~Saenko, ``Deep coral: Correlation alignment for deep domain adaptation,'' in \emph{Computer Vision--ECCV 2016 Workshops: Amsterdam, The Netherlands, October 8-10 and 15-16, 2016, Proceedings, Part III 14}.\hskip 1em plus 0.5em minus 0.4em\relax Springer, 2016, pp. 443--450.

\bibitem{zou2018unsupervised}
Y.~Zou, Z.~Yu, B.~Kumar, and J.~Wang, ``Unsupervised domain adaptation for semantic segmentation via class-balanced self-training,'' in \emph{Proceedings of the European conference on computer vision (ECCV)}, 2018, pp. 289--305.

\bibitem{du2022learning}
Y.~Du, Y.~Shen, H.~Wang, J.~Fei, W.~Li, L.~Wu, R.~Zhao, Z.~Fu, and Q.~Liu, ``Learning from future: A novel self-training framework for semantic segmentation,'' \emph{Advances in Neural Information Processing Systems}, vol.~35, pp. 4749--4761, 2022.

\bibitem{zheng2021rectifying}
Z.~Zheng and Y.~Yang, ``Rectifying pseudo label learning via uncertainty estimation for domain adaptive semantic segmentation,'' \emph{International Journal of Computer Vision}, vol. 129, no.~4, pp. 1106--1120, 2021.

\bibitem{sohn2020fixmatch}
K.~Sohn, D.~Berthelot, N.~Carlini, Z.~Zhang, H.~Zhang, C.~A. Raffel, E.~D. Cubuk, A.~Kurakin, and C.-L. Li, ``Fixmatch: Simplifying semi-supervised learning with consistency and confidence,'' \emph{Advances in neural information processing systems}, vol.~33, pp. 596--608, 2020.

\bibitem{zhou2022context}
Q.~Zhou, Z.~Feng, Q.~Gu, J.~Pang, G.~Cheng, X.~Lu, J.~Shi, and L.~Ma, ``Context-aware mixup for domain adaptive semantic segmentation,'' \emph{IEEE Transactions on Circuits and Systems for Video Technology}, vol.~33, no.~2, pp. 804--817, 2022.

\bibitem{chopra2005learning}
S.~Chopra, R.~Hadsell, and Y.~LeCun, ``Learning a similarity metric discriminatively, with application to face verification,'' in \emph{2005 IEEE computer society conference on computer vision and pattern recognition (CVPR'05)}, vol.~1.\hskip 1em plus 0.5em minus 0.4em\relax IEEE, 2005, pp. 539--546.

\bibitem{zhao2021contrastive}
X.~Zhao, R.~Vemulapalli, P.~A. Mansfield, B.~Gong, B.~Green, L.~Shapira, and Y.~Wu, ``Contrastive learning for label efficient semantic segmentation,'' in \emph{Proceedings of the IEEE/CVF International Conference on Computer Vision}, 2021, pp. 10\,623--10\,633.

\bibitem{hu2021region}
H.~Hu, J.~Cui, and L.~Wang, ``Region-aware contrastive learning for semantic segmentation,'' in \emph{Proceedings of the IEEE/CVF International Conference on Computer Vision}, 2021, pp. 16\,291--16\,301.

\bibitem{li2021semantic}
S.~Li, B.~Xie, B.~Zang, C.~H. Liu, X.~Cheng, R.~Yang, and G.~Wang, ``Semantic distribution-aware contrastive adaptation for semantic segmentation,'' \emph{arXiv preprint arXiv:2105.05013}, 2021.

\bibitem{khosla2020supervised}
P.~Khosla, P.~Teterwak, C.~Wang, A.~Sarna, Y.~Tian, P.~Isola, A.~Maschinot, C.~Liu, and D.~Krishnan, ``Supervised contrastive learning,'' \emph{Advances in neural information processing systems}, vol.~33, pp. 18\,661--18\,673, 2020.

\bibitem{he2020momentum}
K.~He, H.~Fan, Y.~Wu, S.~Xie, and R.~Girshick, ``Momentum contrast for unsupervised visual representation learning,'' in \emph{Proceedings of the IEEE/CVF conference on computer vision and pattern recognition}, 2020, pp. 9729--9738.

\bibitem{vayyat2022cluda}
M.~Vayyat, J.~Kasi, A.~Bhattacharya, S.~Ahmed, and R.~Tallamraju, ``Cluda: Contrastive learning in unsupervised domain adaptation for semantic segmentation,'' \emph{arXiv preprint arXiv:2208.14227}, 2022.

\bibitem{brown2020language}
T.~Brown, B.~Mann, N.~Ryder, M.~Subbiah, J.~D. Kaplan, P.~Dhariwal, A.~Neelakantan, P.~Shyam, G.~Sastry, A.~Askell \emph{et~al.}, ``Language models are few-shot learners,'' \emph{Advances in neural information processing systems}, vol.~33, pp. 1877--1901, 2020.

\bibitem{bao2021beit}
H.~Bao, L.~Dong, S.~Piao, and F.~Wei, ``Beit: Bert pre-training of image transformers,'' \emph{arXiv preprint arXiv:2106.08254}, 2021.

\bibitem{dosovitskiy2020image}
A.~Dosovitskiy, L.~Beyer, A.~Kolesnikov, D.~Weissenborn, X.~Zhai, T.~Unterthiner, M.~Dehghani, M.~Minderer, G.~Heigold, S.~Gelly \emph{et~al.}, ``An image is worth 16x16 words: Transformers for image recognition at scale,'' \emph{arXiv preprint arXiv:2010.11929}, 2020.

\bibitem{li2022mc}
X.~Li, Y.~Ge, K.~Yi, Z.~Hu, Y.~Shan, and L.-Y. Duan, ``mc-beit: Multi-choice discretization for image bert pre-training,'' in \emph{European Conference on Computer Vision}.\hskip 1em plus 0.5em minus 0.4em\relax Springer, 2022, pp. 231--246.

\bibitem{wei2022masked}
C.~Wei, H.~Fan, S.~Xie, C.-Y. Wu, A.~Yuille, and C.~Feichtenhofer, ``Masked feature prediction for self-supervised visual pre-training,'' in \emph{Proceedings of the IEEE/CVF Conference on Computer Vision and Pattern Recognition}, 2022, pp. 14\,668--14\,678.

\bibitem{he2022masked}
K.~He, X.~Chen, S.~Xie, Y.~Li, P.~Doll{\'a}r, and R.~Girshick, ``Masked autoencoders are scalable vision learners,'' in \emph{Proceedings of the IEEE/CVF conference on computer vision and pattern recognition}, 2022, pp. 16\,000--16\,009.

\bibitem{kakogeorgiou2022hide}
I.~Kakogeorgiou, S.~Gidaris, B.~Psomas, Y.~Avrithis, A.~Bursuc, K.~Karantzalos, and N.~Komodakis, ``What to hide from your students: Attention-guided masked image modeling,'' in \emph{European Conference on Computer Vision}.\hskip 1em plus 0.5em minus 0.4em\relax Springer, 2022, pp. 300--318.

\bibitem{pathak2016context}
D.~Pathak, P.~Krahenbuhl, J.~Donahue, T.~Darrell, and A.~A. Efros, ``Context encoders: Feature learning by inpainting,'' in \emph{Proceedings of the IEEE conference on computer vision and pattern recognition}, 2016, pp. 2536--2544.

\bibitem{xie2022simmim}
Z.~Xie, Z.~Zhang, Y.~Cao, Y.~Lin, J.~Bao, Z.~Yao, Q.~Dai, and H.~Hu, ``Simmim: A simple framework for masked image modeling,'' in \emph{Proceedings of the IEEE/CVF conference on computer vision and pattern recognition}, 2022, pp. 9653--9663.

\bibitem{xuestare}
H.~Xue \emph{et~al.}, ``Stare at what you see: Masked image modeling without reconstruction.” arxiv, nov. 16, 2022. accessed: Nov. 30, 2022.''

\bibitem{yang2020fda}
Y.~Yang and S.~Soatto, ``Fda: Fourier domain adaptation for semantic segmentation,'' in \emph{Proceedings of the IEEE/CVF conference on computer vision and pattern recognition}, 2020, pp. 4085--4095.

\bibitem{lv2020cross}
F.~Lv, T.~Liang, X.~Chen, and G.~Lin, ``Cross-domain semantic segmentation via domain-invariant interactive relation transfer,'' in \emph{Proceedings of the IEEE/CVF Conference on Computer Vision and Pattern Recognition}, 2020, pp. 4334--4343.

\bibitem{mei2020instance}
K.~Mei, C.~Zhu, J.~Zou, and S.~Zhang, ``Instance adaptive self-training for unsupervised domain adaptation,'' in \emph{Computer Vision--ECCV 2020: 16th European Conference, Glasgow, UK, August 23--28, 2020, Proceedings, Part XXVI 16}.\hskip 1em plus 0.5em minus 0.4em\relax Springer, 2020, pp. 415--430.

\bibitem{tranheden2021dacs}
W.~Tranheden, V.~Olsson, J.~Pinto, and L.~Svensson, ``Dacs: Domain adaptation via cross-domain mixed sampling,'' in \emph{Proceedings of the IEEE/CVF Winter Conference on Applications of Computer Vision}, 2021, pp. 1379--1389.

\bibitem{wang2021domain}
Q.~Wang, D.~Dai, L.~Hoyer, L.~Van~Gool, and O.~Fink, ``Domain adaptive semantic segmentation with self-supervised depth estimation,'' in \emph{Proceedings of the IEEE/CVF International Conference on Computer Vision}, 2021, pp. 8515--8525.

\bibitem{zhang2021prototypical}
P.~Zhang, B.~Zhang, T.~Zhang, D.~Chen, Y.~Wang, and F.~Wen, ``Prototypical pseudo label denoising and target structure learning for domain adaptive semantic segmentation,'' in \emph{Proceedings of the IEEE/CVF conference on computer vision and pattern recognition}, 2021, pp. 12\,414--12\,424.

\bibitem{chen2023ida}
Z.~Chen, Z.~Ding, J.~M. Gregory, and L.~Liu, ``Ida: Informed domain adaptive semantic segmentation,'' \emph{arXiv preprint arXiv:2303.02741}, 2023.

\bibitem{vu2019advent}
T.-H. Vu, H.~Jain, M.~Bucher, M.~Cord, and P.~P{\'e}rez, ``Advent: Adversarial entropy minimization for domain adaptation in semantic segmentation,'' in \emph{Proceedings of the IEEE/CVF conference on computer vision and pattern recognition}, 2019, pp. 2517--2526.

\bibitem{pushp2023povnav}
D.~Pushp, Z.~Chen, C.~Luo, J.~M. Gregory, and L.~Liu, ``Povnav: A pareto-optimal mapless visual navigator,'' \emph{The 18th International Symposium on Experimental Robotics (ISER)}, 2023.

\end{thebibliography}
}
\end{document}